\documentclass{llncs}
\usepackage[american]{babel}
\usepackage[T1]{fontenc}
\usepackage{times}
\usepackage{graphicx}

\begin{document}

\title{Including Dialects and Language Varieties in Author Profiling}
\subtitle{Notebook for PAN at CLEF 2017}

\author{Alina Maria Ciobanu\textsuperscript{1}, Marcos Zampieri\textsuperscript{2}, Shervin Malmasi\textsuperscript{3}, Liviu P. Dinu\textsuperscript{1}}
\institute{\textsuperscript{1}University of Bucharest, Romania \\
\textsuperscript{2}University of Cologne, Germany \\
\textsuperscript{3}Harvard Medical School, USA \\
\texttt{alina.ciobanu@my.fmi.unibuc.ro}}

\maketitle

\begin{abstract}
This paper presents a computational approach to author profiling taking gender and language variety into account. We apply an ensemble system with the output of multiple linear SVM classifiers trained on character and word $n$-grams. We evaluate the system using the dataset provided by the organizers of the 2017 PAN lab on author profiling. Our approach achieved 75\% average accuracy on gender identification on tweets written in four languages and 97\% accuracy on language variety identification for Portuguese.
\end{abstract}

\section{Introduction}

With vast amounts of texts available on social media, author (or authorship) profiling has become a popular research area in NLP. A number of characteristics such as age \cite{nguyen2013}, gender \cite{nguyen2014}, and native language \cite{gebreetal2013,malmasi-cahill2015} can be predicted based on the topics and the linguistic properties present in a person's writings. 

The PAN labs\footnote{\url{http://pan.webis.de/}} at CLEF have been providing a forum for scholars to evaluate authorship profiling approaches on user-generated content. Author profiling tasks organized in the past PAN labs included age, gender, and personality traits prediction \cite{rangel2015overview,rangel2016overview}. This year, for the first time PAN includes language varieties and dialects from four languages: Arabic, English, Portuguese, and Spanish along with gender identification.\footnote{In this paper we make a terminological distinction between (standard national) language varieties and dialects. We consider English, Spanish, and Portuguese to be pluricentric languages each of them including their own standard national language varieties. The situation of Arabic is, however, different as Modern Standard Arabic (MSA) co-exists with several Arabic dialects in a diglossic situation. Nevertheless, the challenges faced by systems trained to discriminate between similar languages, language varieties, and dialects are identical.}

This paper describes computational methods for gender and language variety identification on social media. Our approach builds on the experience acquired in the previous gender identification tasks of the PAN labs and the four editions of the Discriminating between Similar Languages (DSL)\footnote{\url{http://ttg.uni-saarland.de/vardial2017/sharedtask2017.html}} shared task organized at the workshop on Similar Languages, Varieties and Dialects (VarDial)  \cite{zampieri:2014:VarDial,zampieri:2015:LT4VarDial,dsl2016,zampieri2017findings}. The DSL shared tasks included all languages\footnote{Arabic dialect identification (ADI) was a sub-task of the DSL 2016 and an individual task in the more comprehensive VarDial evaluation campaign 2017.} and most of the dialects and language varieties included in the PAN lab 2017 thus establishing benchmarks for language variety and dialect identification.

\section{Related Work}

The inclusion of language varieties at PAN is motivated by the growing interest in dialect and language variety identification evidenced by several research papers and the aforementioned DSL and ADI shared tasks. Examples of such studies include Portuguese varieties \cite{zampieriandgebre12,zampieri2016,castro2017smoothed}, English varieties \cite{lui13}, Romanian dialects \cite{ciobanu2016}, Chinese varieties \cite{xu2016sentence}, and a number of studies on Arabic dialect identification \cite{tillmann-mansour-alonaizan:2014:VarDial,zaidan14,sadat14,malmasi-et-al:2015:adi}.

The DSL and ADI shared task reports and their respective system description papers provide valuable information about successful approaches for dialect, language variety, and similar language identification. Successful approaches such as those by Goutte et al. (2014) \cite{goutteetal2014}, Malmasi and Dras (2015) \cite{malmasi2015dsl}, Malmasi and Zampieri (2016) \cite{dsl2016malmasi}, and Bestgen (2017) \cite{bestgen2017improving} rely on the combination of higher-order character $n$-grams (4 and above), word $n$-grams, POS tags in \cite{bestgen2017dsl}, and multiple linear classifiers such as SVMs and Naive Bayes arranged in ensembles and/or trained in a two-stage approach, in which first the language is identified and subsequently individual classifiers are trained to discriminate between language varieties or dialects of the same language.\footnote{Goutte et al. (2016) \cite{dslrec:2016} provides a comprehensive evaluation of the first two editions of the DSL shared task.} An exception is the approach proposed by Ionescu and Butnaru (2017) \cite{ionescu2017dsl} which achieved great results for Arabic dialect identification relying on kernel learning.

The main difference between the language variety sub-task at PAN and the DSL and ADI shared tasks is the kind of data provided by the organizers. The PAN challenge provides data collected from social media, whereas the data used in the DSL task comes from newspapers \cite{tan14} and the ADI shared tasks used transcripts from broadcast speeches along with audio features \cite{Ali+2016}. With respect to the data, the most similar task to the PAN challenge is the 2014 TweetLID shared task \cite{zubiaga14} which included microblog posts from the languages spoken in the Iberian Peninsula and English. 

\section{Methods}

\subsection{Task and Data}

The organizers of the PAN challenge on author profiling provided participants with a training set containing \textasciitilde1,140,000 microblog posts from {\em Twitter}. Each post in the training set was annotated with the user's metadata including the language, language variety or dialect, and gender. A test set including unlabeled posts was released a month later.

The four languages and their respective varieties and dialects included in the PAN 2017 dataset are listed next.

\begin{itemize}
\item Arabic: Egypt, Gulf, Levantine, Maghrebi.
\item English: Australia, Canada, Great Britain, Ireland, New Zealand, United States.
\item Portuguese: Brazil, Portugal.
\item Spanish: Argentina, Chile, Colombia, Mexico, Peru, Spain, Venezuela.
\end{itemize}

\noindent The training set was annotated in XML format. Next we present an example of the meta-data for a male English speaker from the United States.

\begin{verbatim}
<author id="author-id" lang="en" variety="united states"
gender="male">
\end{verbatim}

\noindent With the data provided by the PAN 2017 organizers in hand we trained SVM classifiers to identify both the gender and the language variety or dialect of users. Participants could choose to participate in any or both sub-tasks and we decided to participate in both.

Finally, it is worth noting that, unlike most NLP shared tasks, PAN requires participants to run their scripts in a virtual machine provided by the organizers. This ensures that all teams have the same computing power to participate in the challenge allowing full reproducibility \cite{stein:2014j}.\footnote{The PAN labs use TIRA (\url{http://www.tira.io/}) for reproducibility.} 

\subsection{Approach}

We use a single-label multi-class classification approach based on SVM ensembles, following the methodology proposed by Malmasi and Dras \cite{malmasi2015dsl}.

Classification ensembles are systems that combine the results of multiple classifiers, with the purpose of improving the overall performance. Ensembles have been successfully used in various research areas, such as complex word identification \cite{malmasi2016ltg} or grammatical error diagnosis \cite{xiang_et_al_2015}. The individual classifiers can differ in various regards, such as training data, features or classification methods.

In our system, the classifiers differ in terms of features. We use character $n$-grams (with $n$ in $\{1, ..., 6\}$) and word $n$-grams (with $n$ in $\{1, 2\}$) and build a classifier for each type of feature. Thus, our ensemble consists of eight individual classifiers. To combine the classifiers, we employ a fusion method based on the probability estimates provided by the individual classifiers: the predicted probabilities for each class are added, and the prediction of the ensemble is the class with the highest sum. We use the SVM implementation provided by Scikit-learn \cite{scikit-learn}, based on the Libsvm library \cite{libsvm}.

We train the ensembles individually for predicting gender and language varieties. We perform 3-fold cross-validation on the training dataset for hyperparameter tuning, for each classifier, searching for the optimal value of $C$ in $\{10^{-5}, ..., 10^5\}$.

\section{Results}

In the next two sections we present the results obtained by our method. Section \ref{sec:cv} presents the results obtained using cross-validation on the training set. Section \ref{sec:test} presents the official results obtained using the PAN author profiling test set released by the shared task organizers over a month after the training set was released.

\subsection{Cross-Validation}
\label{sec:cv}

The cross-validation results are reported in Table \ref{table:cv} with the best results presented in bold. We note that the highest joint accuracy (when both the gender and language variety are correctly predicted together) is obtained for Portuguese, where the system obtains 0.75 accuracy. For gender identification, the highest accuracy of 0.79 is obtained for English, while language variety is best predicted for Portuguese, with 0.97 accuracy. Portuguese also obtains the highest average accuracy of 0.83 (average of gender, language variety and joint accuracy).

\renewcommand{\tabcolsep}{1.25em}
\renewcommand*{\arraystretch}{1.25}
\begin{table}[!ht]
\caption{\label{table:cv}Cross-validation accuracy on the training set for gender and language variety.}
\begin{center}
\begin{tabular}{l c c c c}
\hline
\textbf{Language} & \textbf{Gender} & \textbf{Variety} & \textbf{Joint} & \textbf{Average} \\
\hline
Arabic & 0.73 & 0.75 & 0.57 & 0.68 \\
English & \bf 0.79 & 0.75 & 0.59 & 0.71 \\
Portuguese & 0.77 & \bf 0.97 & \bf 0.75 & \bf 0.83 \\
Spanish & 0.71 & 0.90 & 0.64 & 0.75 \\
\hline
\end{tabular}
\end{center}
\end{table}

\noindent The high results obtained for Portuguese were not surprising, as there were only two Portuguese varieties in the dataset, from Brazil and from Portugal. The dataset included more varieties and dialects from the other four languages, namely: six English varieties, seven Spanish varieties, and  four Arabic dialects.

The individual classifiers do not outperform, in any case, the ensembles. Portuguese is the only language for which the best individual performance equals the performance of the ensembles. For the others, the improvement reaches a maximum of 0.08 in accuracy (for the English joint prediction) when using ensembles. For three languages out of four (English, Spanish and Portuguese), word unigrams obtain the highest joint accuracy from all the individual classifiers. For Arabic, character 4-grams obtain the highest joint accuracy. As far as the language variety and gender labels are concerned, character 4-grams, character 5-grams and word unigrams obtain better results than the other types of features. For both gender and language variety identification, the best results are obtained for Portuguese, using character 4-grams for gender identification and word unigrams for language variety identification.

\subsection{Test Set}
\label{sec:test}

In the official evaluation carried out on the test set by the PAN organizers our system was ranked 13\textsuperscript{th} among 22 participating teams in both sub-tasks. The system achieved 0.7842 average average accuracy for language variety and gender identification. The results and ranks are described in more detail in the PAN labs report \cite{potthast:2017a} and in the author profiling task report \cite{rangel:2017}.

In Table \ref{table:testvar} we present the results obtained for language variety identification. For reference we provide two baselines provided by the organizers: the BOW-baseline, a bag-of-words model with the 1,000 most frequent items and the STAT-baseline, a simple majority class baseline. As observed in the cross-validation experiments, the best results in the test set were also obtained when discriminating between the two Portuguese varieties achieving 0.9788 accuracy. On language variety identification our system achieved an average performance of 0.8524 accuracy ranking 11\textsuperscript{th} among 22 shared task entries.

\renewcommand{\tabcolsep}{1.25em}
\renewcommand*{\arraystretch}{1.25}
\begin{table}[!ht]
\caption{\label{table:testvar}Test set accuracy results for language variety identification.}
\begin{center}
\begin{tabular}{c c c c c c}
\hline
\textbf{Rank} & \textbf{Arabic} & \textbf{English} & \textbf{Portuguese} & \textbf{Spanish} & \textbf{Average} \\
\hline
11\textsuperscript{th} of 22 &	0.7569 &	0.7746 & 0.9788	& 0.8993 &	0.8524 \\
BOW-baseline &	0.3394 &	0.6592 &	0.9712 & 0.7929 &	0.6907 \\
STAT-baseline &	0.2500 &	0.1667 &	0.5000 &	0.1429 &	0.2649 \\
\hline
\end{tabular}
\end{center}
\end{table}

\noindent In Table \ref{table:testgender} we present the results obtained for gender identification with tweets from different languages along with the two aforementioned baselines. This is a binary classification setting in which the systems are trained to discriminate between tweets written by male and female writers. The variable gender was constant between all languages whereas the number of varieties and dialects for each language varied between 2 for Portuguese and 7 for Spanish. For this reason we observed that the results across languages for gender identification varied much less than the results obtained on language variety/dialect identification. 

\renewcommand{\tabcolsep}{1.25em}
\renewcommand*{\arraystretch}{1.25}
\begin{table}[!ht]
\caption{\label{table:testgender}Test set accuracy results for gender identification per language.}
\begin{center}
\begin{tabular}{c c c c c c}
\hline
\textbf{Rank} & \textbf{Arabic} & \textbf{English} & \textbf{Portuguese} & \textbf{Spanish} & \textbf{Average} \\
\hline
12\textsuperscript{th} of 22 &	0.7131 &	0.7642 & 0.7713 &	0.7529 &	0.7504 \\
BOW-baseline &	0.5300 &	0.7075 &	0.7812 &	0.6864 &	0.6763 \\
STAT-baseline &	0.5000 & 0.5000 & 0.5000 & 0.5000 & 0.5000 \\
\hline
\end{tabular}
\end{center}
\end{table}

Our method obtained the best results for Portuguese tweets achieving 0.7713 and the lowest results for Arabic achieving 0.7131 accuracy. The average performance of our method on gender identification was 0.7504 accuracy ranking 12\textsuperscript{th} among 22 shared task entries.

The results presented in this section indicate that our approach performed substantially better than the two baselines provided and it was consistently ranked in the middle of the table both for language variety and for gender identification. Even though the results obtained by our method were not low, taking the experience obtained in the past PAN labs and DSL shared tasks into account we expected our system to rank higher in the official scores table. Possible factors that may have influenced the performance of our method are: 1) the type of dataset used at PAN which contain very short and non-standard texts, 2) the large size of the dataset that might have made possible for the other teams to use innovative approaches (e.g. deep learning), and 3) our implementation of the classifier which might not have been optimal. A thorough analysis of the misclassified instances is being carried out to determine the reasons for this outcome and possible ways to improve our system's performance.

\section{Conclusion}

This paper presented an SVM ensemble-based system trained on character and word $n$-grams developed for author profiling tested on the PAN 2017 dataset which takes gender and language variety/dialect identification into account. The approach described in our submission was inspired by successful submissions to past editions of the PAN task on gender identification, to the Discriminating between Similar Languages (DSL), and to Arabic Dialect Identification (ADI) shared tasks, the last two organized at the VarDial workshop.

In the training set cross-validation stage, our best results for gender identification were obtained on English data, 0.79 accuracy, and the best results for language variety identification were obtained for Portuguese, 0.97 accuracy. In the official evaluation carried out on the test set our system was ranked 11\textsuperscript{th} on language variety identification and 12\textsuperscript{th} on gender identification out of 22 submissions achieving 0.85 and 0.75 accuracy respectively.

To the best of our knowledge, the PAN labs 2017 was the first shared task to include language varieties and dialects in author profiling opening avenues for future research. Regarding our system's performance, there is still room for improvement. We are currently investigating ways to improve our system's performance by testing a meta-classifier which achieved very good results on German dialect identification \cite{malmasi2017bdsl}.

\section*{Acknowledgement}

We would like to thank the organizers of the PAN lab for proposing this interesting shared task. Special thanks to Martin Potthast and Francisco Rangel for replying promptly to all our inquiries and to Paolo Rosso for fruitful discussions and interesting insights about author profiling during the last VarDial workshop at EACL 2017.

Liviu P. Dinu is supported by UEFISCDI, project number 53BG/2016.

\bibliographystyle{splncs03}
\begin{raggedright}
\bibliography{bib}
\end{raggedright}

\end{document}